\pdfoutput=1

\documentclass[11pt]{article}

\usepackage[final]{acl}

\usepackage{times}
\usepackage{latexsym}

\usepackage[T1]{fontenc}

\usepackage[utf8]{inputenc}

\usepackage{microtype}

\usepackage{inconsolata}

\usepackage{graphicx}
\usepackage{float} 
\usepackage{xcolor}
\definecolor{darkgreen}{RGB}{0,200,0}

\usepackage{booktabs}
\usepackage{multirow}
\usepackage{makecell}
\usepackage{tabularx}
\usepackage{arydshln}

\usepackage{amsmath}
\usepackage{amssymb}
\usepackage{amsfonts}
\usepackage[most]{tcolorbox}
\usepackage{tikz}

\newcommand{\modelname}{\textit{MarginSel} }

\usetikzlibrary{shadows}

\usepackage{algorithm}
\usepackage[noend]{algpseudocode}

\usepackage{caption}
\usepackage{ulem}
\usepackage{todonotes}

\title{\modelname: Max-Margin Demonstration Selection for LLMs}

\author{Rajeev Bhatt Ambati \\
  UNC Chapel Hill \\
  \texttt{ambati@cs.unc.edu} \\\And
  James Lester \\
  NC State University \\
  \texttt{lester@ncsu.edu} \\\AND
  Shashank Srivastava \\
  UNC Chapel Hill \\
  \texttt{ssrivastava@cs.unc.edu} \\\And
  Snigdha Chaturvedi \\
  UNC Chapel Hill \\
  \texttt{snigdha@cs.unc.edu} \\}

\begin{document}
\maketitle
\begin{abstract}
Large Language Models (LLMs) excel at few-shot learning via in-context learning (ICL). However, the effectiveness of ICL is often sensitive to the selection and ordering of demonstration examples \cite{order_sensitivity, good_examples_gpt3}. To address this, we present \modelname: Max-\uline{Margin} Demonstration \uline{Sel}ection for LLMs, a two-step method that selects hard demonstration examples for the ICL prompt, adapting to each test instance. Our approach achieves 2-7\% absolute improvement in $F_{1}$-score across classification tasks, compared to a random selection of examples. We also provide theoretical insights and empirical evidence showing that \modelname induces max-margin behavior in LLMs by effectively increasing the margin for hard examples—analogous to support vectors—thereby shifting the decision boundary in a beneficial direction.
\end{abstract}

\section{Introduction}
\label{sec:intro}

Large Language Models (LLMs) \cite{gpt3_paper} perform well on a wide variety of tasks through few-shot learning via in-context learning (ICL), where task demonstrations are provided within the prompt. However, ICL’s effectiveness is highly sensitive to the selection of demonstrations \cite{order_sensitivity, good_examples_gpt3}. Prior work addresses this by training explicit retrievers \cite{lu2023dynamicpromptlearningpolicyPGPrompt, wang2024demonstrationselectionincontextlearningRDES}, which optimize demonstration selection using labeled data. However, these approaches are computationally costly and task-specific, requiring retraining for each new task or LLM.  Some methods avoid training by relying on semantic similarity, yet they neglect the LLM's inherent uncertainty near the decision boundary. This is a critical factor for tasks with fine-grained distinctions between classes.

\begin{figure}[t]
  \includegraphics[width=\columnwidth]{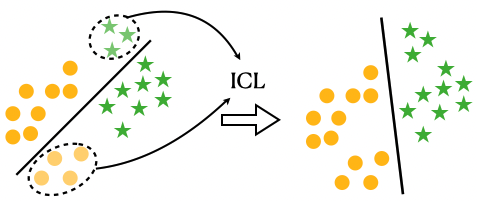}
  \caption{\modelname identifies misclassified examples near the decision boundary (circled) for a two-class classification task (yellow circles and green stars). By including these high-margin examples in the ICL prompt, \modelname shifts the decision boundary, improving predictions and inducing max-margin behavior in LLMs.}
  \label{fig:decision_boundary}
\end{figure}

To address these gaps, we propose \textbf{\modelname: Max-Margin Demonstration Selection}, a novel method that dynamically selects demonstrations near the decision boundary to maximize the classifier margin and adapt to each test instance's uncertainty. \modelname operates in two steps: (1) an LLM is prompted to generate relevant candidate labels for all training and test instances, identifying regions of ambiguity; and (2) training examples with the same candidate labels as the test instance are selected as demonstrations. Drawing inspiration from Support Vector Machines, we posit that these carefully chosen examples act as analogous to \textit{support vectors}, inducing \textit{max-margin behavior} within the LLM. As illustrated in Figure~\ref{fig:decision_boundary}, incorporating ambiguous examples(dotted), which are likely to be boundary-proximal, drives a beneficial shift in the decision boundary. 

Empirical evaluations across four LLMs and three tasks (e.g., cognitive distortion detection, sentiment classification) show that \modelname consistently outperforms standard prompting and kNN-ICL\cite{good_examples_gpt3}, achieving absolute F1-score gains of 2–7\%. Furthermore, our theoretical analysis demonstrates that \modelname examples induce a max-margin behavior within LLMs, a finding corroborated by empirical analysis showing significantly increased class separation compared to random selection.

Our key contributions are:
\begin{itemize}
    \item \textbf{\modelname: Max-Margin Demonstration Selection}, a novel method that self-guides demonstration selection using the LLM’s uncertainty signals.
    \item \textbf{Theoretical Foundation:} A formal connection between \modelname and max-margin optimization, showing its equivalence to SVM-like boundary adjustment in LLMs.
    \item \textbf{Empirical Validation:} Robust improvements across model scales and tasks, with ablations highlighting the critical role of boundary-aware examples.
\end{itemize}

\section{Related Work}
\label{sec:related_work}

While in-context learning (ICL) has enabled LLMs to solve complex tasks using only a few examples \cite{gpt3_paper}, the choice and ordering of demonstration examples critically affect performance. For instance, \citet{order_sensitivity} showed that permuting the order of demonstrations can lead to significant output variations, underscoring the need for robust demonstration selection. Existing approaches for demonstration selection broadly fall into two categories: \textbf{retriever-trained} methods that optimize selection via learned models and \textbf{retriever-free} methods.

\noindent \textbf{Retriever-trained methods} optimize example selection using task-specific labels. For example, \citet{rubin-etal-2022-retriever} trains a retriever using contrastive learning to score input-output pairs, while \citet{wang2024demonstrationselectionincontextlearningRDES} uses a deep Q-network to balance label diversity and relevance. Similarly, \citet{lu2023dynamicpromptlearningpolicyPGPrompt} trains a reinforcement learning (RL) agent to optimize demonstration ordering for downstream task accuracy. Although effective, these methods incur high computational costs and require retraining for each task or LLM, limiting their scalability.

\noindent \textbf{Retriever-free methods} avoid retriever training. A widely adopted approach, kNN-ICL \cite{good_examples_gpt3}, retrieves examples based on semantic similarity in an embedding space (e.g., RoBERTa \cite{liu2020roberta}). Other works, such as \citet{levy2023diversedemonstrationsimproveincontext}, prioritize diversity through clustering or determinantal point processes. While efficient, these methods fail to adapt to task-specific ambiguities. For instance, kNN-ICL often selects redundant or easily classified examples, providing limited signal for refining decision boundaries near regions of uncertainty.

Our work \modelname is also a retriever-free method that dynamically adapts demonstrations to each test instance’s ambiguity pattern. Unlike retriever-trained methods, \modelname requires no task-specific optimization, and unlike existing retriever-free methods like kNN-ICL, it self-guides selection using the LLM’s uncertainty signals. It selects demonstration examples near the decision boundary that act as \textit{support vectors} (Section~\ref{sec:hard_examples_help}). \modelname thus bridges the gap between these two categories, offering both efficiency and adaptability. In our experiments, we compare with kNN-ICL, as it is a widely adopted retriever-free method and the most relevant baseline for our approach. Theoretically, we extend \cite{icl_meta_grad_theory}'s duality between ICL and gradient descent to show that \modelname induces \textit{max-margin behavior} (Section~\ref{sec:other_examples_help}), a property unaddressed by prior approaches.

\section{Max-Margin Demonstration Selection}
\label{sec:method}
For a classification task, given a test example\footnote{$(.)^{t}$ denotes a variable whose datatype is token(s)} $x^t_{\text{test}}$ (a sequence of tokens) and a label set $\{y^t_c\}_{c=1}^{C}$, where $C$ is the number of classes, the goal is to predict a label $\hat{y}^t$.
In zero-shot learning, we predict
\[
\hat{y}^t = \arg\max_{y^t_c} P(y^t_c \mid x^t_{\text{test}}, \mathcal{M}),
\]
for a given model $\mathcal{M}$. In in-context learning (ICL) using a set of demonstration examples
\[
\mathcal{D} = \{(x^t_{\text{demo}, k}, y^t_{\text{demo}, k})\}_{k=1}^{N}
\]
where each $x^t_{\text{demo}, k}$ is the demonstration input and $y^t_{\text{demo}, k}$ is the corresponding demonstration label, we predict
\[
\hat{y}^t = \arg\max_{y^t_c} P(y^t_c \mid x^t_{\text{test}}, \mathcal{D}, \mathcal{M}).
\]
The goal of \modelname is to identify the demonstration set $\mathcal{D}$ such that it leads to the most accurate prediction $\hat{y}^t$ for a test example $x^t_{\text{test}}$.

\begin{figure*}[t]
    \centering
    \includegraphics[width=0.8\linewidth]{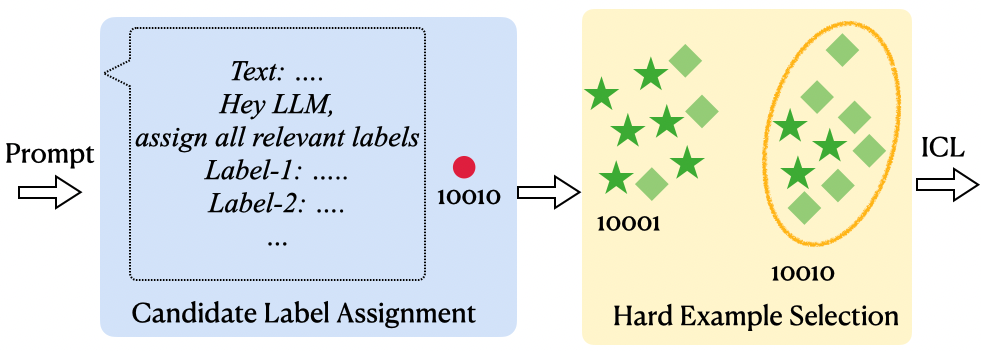}
    \caption{\textit{\modelname}: In Step 1 (Candidate Label Assignment), the LLM is prompted to assign all relevant candidate labels, which are displayed in binary form (e.g., 10010 indicates that labels 1 and 4 are assigned). In Step 2 (Hard Example Selection), training examples with matching candidate labels (10010) are selected for ICL—these examples may belong to multiple classes (depicted by stars and diamonds are 2 out of 5 classes).}
    \label{fig:method}
\end{figure*}

\modelname draws motivation from hard mining \cite{hard_mining}, where misclassified examples are identified and incorporated into the training set. In hard mining, the gradients computed from examples near the decision boundary help update the model's weights to correct the boundary. Inspired by this idea, \modelname first identifies hard examples by prompting the LLM to assign all relevant candidate labels in a zero-shot manner (see Figure~\ref{fig:method}) and then selects training examples that have the same candidate labels. Our approach consists of two steps:

\begin{algorithm}
\caption{$\alpha$-\modelname}
\begin{algorithmic}[1]
\State \textbf{Inputs:}

\begin{itemize}
    \item Training examples \(\mathcal{E} = \{(x^t_i, y^t_i)\}_{i=1}^{N}\)
    \item Test sample \(x^t_{\text{test}}\)
    \item Zero-shot multilabel classifier \(C_{\text{ZS}}(\cdot)\)
    \item Number of samples \(n\) and ratio \(\alpha\)
    \item Label frequency function \(\rho(y^t)\) (the proportion of samples with label \(y^t\) in \(\mathcal{E}\))
\end{itemize}

\State \textbf{Outputs:}
\begin{itemize}
    \item Demonstration set \(\mathcal{S}\) for test sample \(x^t_{\text{test}}\)
\end{itemize}

\State \textbf{Step 1: Candidate Label Assignment}
\State \(\mathcal{L} \gets \varnothing\) \hfill \(\triangleright\) initialize lookup table
\For{each \((x^t_i, y^t_i) \in \mathcal{E}\)} \hfill \(\triangleright\) training set
    \State \(C_{\text{cand}, i} \gets C_{\text{ZS}}(x^t_i)\)
    \State \(\mathcal{L} \gets \mathcal{L} \cup \{(x^t_i, y^t_i, C_{\text{cand}, i})\}\) \hfill \(\triangleright\) update
\EndFor
\State \(C_{\text{cand}, \text{test}} \gets C_{\text{ZS}}(x^t_{\text{test}})\) \hfill \(\triangleright\) test example

\State \textbf{Step 2: Hard Example Selection}
\State \(\mathcal{S}_{\text{\modelname}} \gets \varnothing\)
\For{each \((x^t_i, y^t_i, C_{\text{cand}, i}) \in \mathcal{L}\)}
    \If{\(C_{\text{cand}, i} = C_{\text{cand}, \text{test}}\)}
        \State \(\mathcal{S}_{\text{\modelname}} \gets \mathcal{S}_{\text{\modelname}} \cup \{(x^t_i, y^t_i)\}\)
    \EndIf
\EndFor
\If{\(|\mathcal{S}_{\text{\modelname}}| > \alpha \cdot n\)}
    \ForAll{\((x^t_i, y^t_i) \in \mathcal{S}_{\text{\modelname}}\)}
        \State \(w_i \gets \frac{1}{\rho(y^t_i)}\) \hfill \(\triangleright\) sampling weight
    \EndFor
    \ForAll{\((x^t_i, y^t_i) \in \mathcal{S}_{\text{\modelname}}\)}
        \State \(\tilde{w}_i = \frac{w_i}{\sum_{(x^t_j, y^t_j) \in \mathcal{S}_{\text{\modelname}}} w_j}\) \hfill \(\triangleright\) normalize
    \EndFor
    \State $\mathcal{S}_{\text{\modelname}} \leftarrow \operatorname{Sample}\Bigl(\mathcal{S}_{\text{\modelname}},\, \alpha \cdot n,\, \{\tilde{w}_i\}\Bigr)$
\EndIf

\State $\mathcal{S}_{\text{kNN}} \leftarrow \operatorname{kNN}\Big((1 - \alpha) \cdot n, \mathcal{E})$ \hfill \(\triangleright\) kNN-ICL
\State $\mathcal{S} = \mathcal{S}_{\text{\modelname}} \cup \mathcal{S}_{\text{kNN}}$
\State \Return $\mathcal{S}$.

\end{algorithmic}
\label{algo:method_algo}
\end{algorithm}

\noindent \textbf{Candidate Label Assignment:} In this step, we prompt the LLM to assign all relevant candidate labels for an example using a zero-shot approach. This is done for the full training set (only once offline) and for the test example $x^t_{\text{test}}$. The underlying intuition is that if the LLM is uncertain about an input, it will output multiple candidate labels. The left box in Figure~\ref{fig:method} illustrates this step. In this example, we consider 5 classes where the LLM assigns candidate labels 1 and 4 (shown as 10010) to the test example (shown as a red circle). This step is presented in lines 3-8 of the Algorithm~\ref{algo:method_algo}.

\noindent \textbf{Hard Example Selection:} Next, we consider training examples that have the same candidate labels determined in Step-1. The right box in Figure~\ref{fig:method} illustrates this step, where the LLM assigns candidate labels 1 and 5 (shown as 10001) to some training examples and candidate labels 1 and 4 (shown as 10010) to others. For the given test example (red circle), \modelname selects the training examples with candidate labels 1 and 4 (inside yellow circle) as demonstrations. From these examples, we sample the required number of shots. However, the distribution of classes in the training data may be imbalanced, leading to a bias in the selection of examples. To account for this imbalance, we perform weighted sampling, where each example is assigned a weight equal to the inverse of its corresponding label frequency. Notice that the selection is based on the candidate labels assigned by the LLM, thereby adapting to a given test example.

In addition, we also incorporate the idea of using semantically similar examples. For this, we also obtain examples using kNN-ICL and include them in the set of demonstration examples. A hyperparameter \(\alpha\) controls the ratio of hard examples to semantically similar examples; specifically, \(\alpha = 0\) implies that only kNN-ICL examples are used, while \(\alpha = 1.0\) means that only hard examples are selected. This step is presented in lines 9-22 of the Algorithm~\ref{algo:method_algo}. The prompts used for the Candidate Label Assignment and Final Label Prediction for all the datasets are provided in Section~\ref{sec:prompts}.

\section{Theoretical Analysis}
\label{sec:theory}
Building on insights from prior work \cite{icl_meta_grad_theory}, we explain why the hard examples selected by \modelname are advantageous, and whether other examples similarly contribute to performance improvements. We show that these examples function analogously to support vectors, in a support vector machine (SVM), effectively increasing the margin and enhancing overall performance.

\subsection{Why Hard Examples Help?}
\label{sec:hard_examples_help}

We adopt a similar formulation as \citet{icl_meta_grad_theory} to motivate our \modelname approach. For ICL, both the demonstration inputs and labels are fed along with the test example (all tokens) in a prompt to the LLM as $[\mathcal{D}; x^t_{\text{test}}; q^t]$. Here, $q^t$ is the query token (e.g., the tokens corresponding to "answer:" or "label:"). Now lets denote the corresponding vector representations as follows: Let
\begin{itemize}
    \item $\mathbf{q} \in \mathbb{R}^{d}$ be the vector representation of the query token in $x_{test}^{t}$.
    \item $\mathbf{X} \in \mathbb{R}^{M \times d}$ be a matrix where each row is the vector representation of all tokens in $x_{test}^{t}$ before the query token.
    \item $\mathbf{X}' = [\mathbf{x}'_{1}, \dots, \mathbf{x}'_{N}]^T \in \mathbb{R}^{N \times d}$ be a matrix where each row $\mathbf{x}'_{k}$ is the vector representation of the demonstration input $x_{\text{demo}, k}^{t}$ from the demonstration set $\mathcal{D}$.
    \item $\mathbf{Y}' = [\mathbf{y}'_{1}, \dots, \mathbf{y}'_{N}]^T \in \mathbb{R}^{N \times d}$ be a matrix where each row $\mathbf{y}'_{k}$ is the vector representation of the demonstration label $y_{\text{demo}, k}^{t}$ from the demonstration set $\mathcal{D}$.
\end{itemize}

Here, $d$ denotes the dimension of the hidden vectors used in the LLM's Transformer architecture. The vector representation of the full prompt fed to the LLM is $[\mathbf{X}'; \mathbf{Y}'; \mathbf{X}]$. Here, $[\,\cdot\,;\,\cdot\,]$ denotes concatenation. Let $\mathcal{F}_{\text{ICL}}$ denote the attention mechanism over this ICL prompt. We compute the context vector obtained by $\mathbf{q}$ via the attention mechanism:
\begin{equation}
\begin{aligned}
\mathcal{F}_{\text{ICL}}(\mathbf{q}) &= \mathrm{Attn}(V, K, \mathbf{q}) \\
&= W_{V} [\mathbf{X}'; \mathbf{Y}'; \mathbf{X}] \\
&\quad \mathrm{softmax}\!\left( \frac{(W_{K} [\mathbf{X}'; \mathbf{Y}'; \mathbf{X}])^T W_{Q} \mathbf{q}}{\sqrt{d}} \right)
\end{aligned}
\end{equation}

where $W_{V}$, $W_{K}$, and $W_{Q} \in \mathbb{R}^{d' \times d}$ are the projection matrices for values, keys, and queries, respectively. For analytical simplicity, consider a relaxed version of attention—linear attention (denoted by $\mathcal{\tilde{F}}_{\text{ICL}}$)—where we omit the softmax function and the scaling factor:
\begin{equation}
\begin{aligned}
\mathcal{\tilde{F}}_{\text{ICL}}(\mathbf{q}) &= \mathrm{LinearAttn}(V, K, \mathbf{q}) \\
&= W_{V} [\mathbf{X}'; \mathbf{Y}'; \mathbf{X}] \\
&\quad (W_{K} [\mathbf{X}'; \mathbf{Y}'; \mathbf{X}])^T W_{Q} \mathbf{q}
\end{aligned}
\end{equation}

This relaxed form can be decomposed into separate contributions:
\begin{equation}
\begin{aligned}
\mathcal{\tilde{F}}_{\text{ICL}}(\mathbf{q}) &= W_{V}\mathbf{X}(W_{K}\mathbf{X})W_{Q}\mathbf{q} \; + \\
&\quad W_{V}\mathbf{X}'(W_{K}\mathbf{X}')W_{Q}\mathbf{q} \; + \\
&\quad W_{V}\mathbf{Y}'(W_{K}\mathbf{Y}')W_{Q}\mathbf{q}
\end{aligned}
\end{equation}

Here, the first term corresponds to the context vector derived from the test input $x^t_{\text{test}}$ (with representation $\mathbf{X}$), the second term to the demonstration inputs $\{x^t_{\text{demo}, k}\}_{k=1}^{N}$ (with representation \(\mathbf{X}'\)), and the third term to the demonstration labels $\{y^t_{\text{demo}, k}\}_{k=1}^{N}$ (with representation \(\mathbf{Y}'\)). We can express this sum as:
\begin{equation}
\begin{aligned}
\tilde{\mathcal{F}}_{\text{ICL}}(\mathbf{q}) &= W_{\text{ZSL}}\mathbf{q} + W_{\text{ZSL}}'\mathbf{q} + W_{L}\mathbf{q},
\end{aligned}
\label{eq:icl_zsl_sum}
\end{equation}
where
\begin{align*}
W_{\text{ZSL}} &= W_{V}\mathbf{X}(W_{K}\mathbf{X})W_{Q} \\
W_{\text{ZSL}}' &= W_{V}\mathbf{X}'(W_{K}\mathbf{X}')W_{Q} \\
W_{\text{L}} &= W_{V}\mathbf{Y}'(W_{K}\mathbf{Y}')W_{Q}
\end{align*}

Here, \(W_{\text{ZSL}}\), \(W_{\text{ZSL}}'\) represents the effective parameters when the LLM is prompted zero-shot with the test example $x^t_{\text{test}}$ and demonstration inputs $\{x^t_{\text{demo}, k}\}_{k=1}^{N}$ respectively. \(W_{\text{L}}\) captures the contribution from the demonstration labels $\{y^t_{\text{demo}, k}\}_{k=1}^{N}$. This formulation illustrates that the overall query representation is composed of the zero-shot components from both the test example and the demonstration inputs, along with the label information. A label is assigned to the query $\mathbf{q}$ based on the proximity of $\tilde{\mathcal{F}}_{\text{ICL}}(\mathbf{q})$ with the label representation (third term). In our formulation, the correct label's contribution (3rd term) is maximized when the test input (first term) and demonstration inputs (second term) are well aligned. In \modelname, this is achieved during the hard example selection step (Step 2), where we ensure that only training examples whose candidate labels match those of the test example are selected.

We can rewrite Equation~\ref{eq:icl_zsl_sum} as an update to \(W_{\text{ZSL}}\) from each demonstration example in $\mathbf{X}'$:
\begin{equation}
\begin{aligned}
\mathcal{\tilde{F}}_{\text{ICL}}(\mathbf{q}) &= W_{\text{ZSL}}\mathbf{q} \; + \\
&\quad \sum_{k=1}^{N}W_{V}[\mathbf{x}'_{k};\mathbf{y}'_{k}](W_{K}[\mathbf{x}'_{k};\mathbf{y}'_{k}])^{T}\mathbf{q} \\
&= (W_{\text{ZSL}} + \Delta W_{\text{ICL}})\mathbf{q}
\end{aligned}
\label{eq:affine_update}
\end{equation}

where each $\mathbf{x}'_{k}$ and $\mathbf{y}'_{k}$ are the vector representations of the $k^{th}$ demonstration input $x^t_{\text{demo}, k}$ and its label $y^t_{\text{demo}, k}$, respectively and $\Delta W_{ICL} = \sum_{k=1}^{N}\, \mathbf{y}'_{k} \, (\mathbf{x}'_{k})^T$ denotes the cumulative update from them. This formulation is analogous to a parameter update in an affine layer, and hence attention with ICL can be viewed as a dual form of an affine layer \cite{icl_meta_grad_theory}. This perspective explains why gradients from misclassified (or hard) examples—those lying near the decision boundary—are particularly important, as they drive the update \(\Delta W_{\text{ICL}}\), ultimately shifting the decision boundary initialized solely by zero-shot learning (see Figure~\ref{fig:decision_boundary}).

\subsection{Do Other Examples Help?}
\label{sec:other_examples_help}
From the previous section, we showed that hard examples—those sampled near the decision boundary—are especially beneficial for in-context learning (ICL). An inquisitive observer might naturally ask whether other, less ambiguous examples also contribute to the prediction. Recall that for a classifier to select the correct label, the predicted probability must exceed $1/C$, where $C$ is the number of classes. In practice, the operating point is typically set higher than $1/C$, effectively imposing a margin on the logits. A similar notion applies to LLMs when sampling a label token from the probability distribution over a vocabulary of subword units.

We can reformulate Equation~\ref{eq:icl_zsl_sum} in a way that mirrors the margin constraints of an SVM. Specifically, we want to maximize the geometric margin over the training set $\mathcal{E}$ as
\begin{equation}
    \max_{\gamma,\,W} \gamma \quad \text{s.t.} \quad y_{i}\left(W^{T}\mathbf{x}'_{i} + b\right) \ge \gamma,
    \label{eq:svm_geo_margin}
\end{equation}
and the functional margin as
\begin{equation}
\begin{aligned}
    \max_{\hat{\gamma},\,W} \frac{\hat{\gamma}}{\| W \|_F} \quad
    \text{s.t.} \quad y_{i}\left(W^{T}\mathbf{x}'_{i} + b\right) \ge \hat{\gamma}
\end{aligned}
\label{eq:svm_func_margin}
\end{equation}

where $\gamma$ is the geometric margin, $W = W'_{ZSL}$, and $b = W_{L}\mathbf{q}$.  Here, we use $\mathbf{x}'_{k}$ and $y_{k}$ to represent the vector representation of a generic $k^{th}$ demonstration input and its corresponding label, respectively, in the context of SVM margin discussion. As in SVMs, the prediction depends primarily on support vector examples; the Lagrangian coefficients $\beta_k$ are non-zero only for these critical samples. This insight allows us to decompose Equation~\ref{eq:icl_zsl_sum} as an update to the zero-shot representation that emphasizes the influence of hard examples:
\begin{equation}
\begin{aligned}
    \mathcal{\tilde{F}}_{ICL}(\mathbf{q}) &= W_{ZSL}\mathbf{q} + W_{L}\mathbf{q} + \\ &\quad \sum_{k=1}^{|\mathcal{S}|}\beta_{k}\mathbf{y}'_{k}(\mathbf{x}'_{k})^T \mathbf{q} \\
    &= W_{ZSL}\mathbf{q} + W_{L}\mathbf{q} + \\ &\quad \mathrm{LinearAttn}(\beta \mathbf{Y}', \mathbf{X}', \mathbf{q}) \\
    &= \left( W_{ZSL} + \Delta W_{ICL} \right)\mathbf{q}
\end{aligned}
\label{eq:support_vector_split}
\end{equation}
where the term $\Delta W_{ICL} = \sum_{k=1}^{|\mathcal{S}|}\beta_{k}\, \mathbf{y}'_{k} \, (\mathbf{x}'_{k})^T$ represents the cumulative update driven solely by the hard examples, chosen as the demonstration set $\mathcal{S}$. This formulation clearly shows that, analogous to SVMs, only the examples near the decision boundary (with non-zero $\beta_k$) significantly affect the final prediction. Consequently, examples that are easily classified (i.e., far from the decision boundary) contribute little to the update, underscoring the value of hard examples in ICL.

\section{Experiments}
In this section, we evaluate the performance of \modelname using several large language models (LLMs): Phi-3-small-8k-Instruct \cite{phi3_paper}, Mistral-7B-Instruct-v0.3 \cite{mistral_paper}, Phi-4 \cite{phi4_paper}, Llama-3.1-8B-Instruct \cite{llama_paper}, and Llama-3.1-70B with 8-bit quantization \cite{llama_paper}. All experiments are repeated over three random seeds, with training and testing splits generated via stratified sampling. We compare \modelname with both standard prompting and the most commonly used retriever-free baseline kNN-ICL \cite{good_examples_gpt3}.

\begin{table*}[t]
  \centering
  \resizebox{\linewidth}{!}{%
  \begin{tabular}{llccccc|ccccc}
    \toprule
    \multirow{2}{*}{Dataset} & \multirow{2}{*}{Method} & \multicolumn{5}{c|}{Phi3-mini-8k} & \multicolumn{5}{c}{Mistral-7B} \\
    \cmidrule(lr){3-7} \cmidrule(lr){8-12}
     &  & 2 & 4 & 6 & 8 & 10 & 2 & 4 & 6 & 8 & 10 \\
    \midrule
    \multirow{4}{*}{\makecell{Cognitive \\ Distortion}} 
      & Random & 29.5 & 28.7 & 27.9 & 28.2 & 28.2 & 26.7 & 29.6 & 25.2 & 25.3 & 24.7 \\
      & kNN-ICL \cite{good_examples_gpt3} & 32.9 & 31.6 & 30.1 & 29.6 & 29.6 & 32.6 & 34.2 & 33.0 & 32.3 & 32.2 \\
      & \modelname ($\alpha=1$) & 30.2 & \textbf{33.4} & \textbf{32.1} & 31.9 & 31.9 & 29.9 & 31.8 & 31.2 & 30.9 & 32.0 \\
      & \modelname ($\alpha=0.9$) & 30.7 & \textbf{33.0} & \textbf{33.1} & \textbf{32.4} & \textbf{32.6} & 31.9 & 33.1 & 32.0 & 32.8 & \textbf{33.7} \\
    \midrule
    \multirow{4}{*}{\makecell{Medical \\ Abstracts}} 
      & Random & 55.9 & 57.7 & 53.8 & 56.8 & 60.0 & 55.2 & 53.8 & 53.6 & 54.7 & 54.6 \\
      & kNN-ICL \cite{good_examples_gpt3} & 61.0 & 60.8 & 60.8 & 60.8 & 60.8 & 61.3 & 60.4 & 60.6 & 59.5 & 60.0 \\
      & \modelname ($\alpha=1$) & 61.6 & 61.4 & 61.4 & 61.5 & 61.6 & 56.6 & 57.7 & 57.4 & 58.3 & 58.6 \\
      & \modelname ($\alpha=0.9$) & 62.0 & \textbf{63.0} & \textbf{64.1} & \textbf{63.0} & 62.1 & 61.9 & 61.4 & 61.4 & \textbf{61.8} & \textbf{61.8} \\
    \midrule
    \multirow{4}{*}{SST-5} 
      & Random & 52.9 & 54.0 & 52.1 & 49.3 & 50.3 & 48.6 & 50.5 & 47.1 & 47.0 & 48.0 \\
      & kNN-ICL \cite{good_examples_gpt3} & 54.9 & \textbf{55.0} & 54.2 & 54.6 & 54.8 & 54.9 & 55.0 & 54.2 & 54.6 & 54.8 \\
      & \modelname ($\alpha=1$) & 53.7 & 53.2 & 54.1 & 52.9 & 52.7 & 54.1 & 54.0 & 53.8 & 53.0 & 52.3 \\
      & \modelname ($\alpha=0.9$) & 53.7 & 52.9 & 52.8 & 53.1 & 54.3 & 51.2 & 53.0 & 54.9 &\textbf{ 55.2} & 55.0 \\
    \bottomrule
  \end{tabular}%
  }
  \caption{$F_{1}(\%)$ performance of Phi3-mini-8k and Mistral-7B with few-shot learning on three datasets. “Random” denotes standard prompting with randomly selected demonstration examples; \(\alpha = 1.0\) indicates that only hard examples are used; \(\alpha = 0.9\) corresponds to a composition of 90\% hard examples and 10\% kNN-ICL. Improvements shown in \textbf{bold} are statistically significant over 3 runs.}
  \label{tab:phi3_mistral7}
\end{table*}

\subsection{Datasets}
We evaluated \modelname on three datasets selected to represent a spectrum of label ambiguity, a key factor in assessing its efficacy. These datasets are: \textbf{SST-5} \cite{sst_dataset}, a sentiment classification dataset with five granular labels; \textbf{Cognitive Distortion} \cite{cogdist_dataset}, a challenging dataset for detecting subtle cognitive distortions in text; and \textbf{Medical Abstracts} \cite{medabs_dataset}, a topic classification dataset with relatively less ambiguity. This selection allows us to examine \modelname's effectiveness across tasks with varying degrees of label complexity. Detailed descriptions of each dataset are provided in Appendix~\ref{app:datasets}.

\subsection{Results}
\label{sec:results}
Tables~\ref{tab:phi3_mistral7} and \ref{tab:phi4_llama8} show results from 4 LLMs. “Random” denotes standard prompting with randomly selected demonstrations. “kNN” is the baseline from \cite{good_examples_gpt3}, implemented with RoBERTa sentence embeddings \cite{liu2020roberta} (equivalent to \modelname with \(\alpha=0\)). “\modelname (\(\alpha=1.0\))” uses only hard examples, while “\modelname (\(\alpha=0.9\))” comprises 90\% hard examples and 10\% kNN-selected examples. Hyperparameter tuning of $\alpha$ is shown in Section~\ref{sec:app:results}.

\begin{table*}[t]
  \centering
  \resizebox{\linewidth}{!}{%
  \begin{tabular}{llccccc|ccccc}
    \toprule
    \multirow{2}{*}{Dataset} & \multirow{2}{*}{Method} & \multicolumn{5}{c|}{Phi-4} & \multicolumn{5}{c}{Llama8B} \\
    \cmidrule(lr){3-7} \cmidrule(lr){8-12}
     &  & 2 & 4 & 6 & 8 & 10 & 2 & 4 & 6 & 8 & 10 \\
    \midrule
    \multirow{4}{*}{\makecell{Cognitive \\ Distortion}} 
      & Random & 34.3 & 32.8 & 32.2 & 30.7 & 30.8 & 26.8 & 26.8 & 24.2 & 22.0 & 22.1 \\
      & kNN-ICL \cite{good_examples_gpt3} & 35.6 & 35.7 & 36.9 & 35.4 & 36.4 & \textbf{29.7} & \textbf{30.3} & 30.2 & 28.9 & 30.8 \\
      & \modelname ($\alpha=1$) & 33.0 & 35.0 & 36.1 & 35.0 & 35.5 & 25.6 & 27.2 & 26.8 & 26.5 & 27.8 \\
      & \modelname ($\alpha=0.1$) & \textbf{38.1} & 36.8 & 34.9 & \textbf{37.7} & 37.2 & 25.6 & 28.7 & \textbf{31.4} & \textbf{30.9} & 28.8 \\
    \midrule
    \multirow{4}{*}{\makecell{Medical \\ Abstracts}} 
      & Random & 63.9 & 62.4 & 62.6 & 62.2 & 61.8 & 62.3 & 59.4 & 58.7 & 58.3 & 58.6 \\
      & kNN-ICL \cite{good_examples_gpt3} & 64.0 & 63.4 & 63.2 & 63.4 & 63.2 & 62.9 & 62.3 & 61.8 & 62.1 & 62.4 \\
      & \modelname ($\alpha=1$) & 63.5 & 63.5 & 63.3 & 62.8 & 62.7 & 62.0 & 60.3 & 60.9 & 60.2 & 59.9 \\
      & \modelname ($\alpha=0.1$) & 64.4 & \textbf{64.0} & \textbf{64.5} & 64.2 & 64.2 & \textbf{64.1} & 63.7 & \textbf{62.1} & 61.5 & 63.8 \\
    \midrule
    \multirow{4}{*}{SST-5} 
      & Random & 55.6 & 56.6 & 56.0 & 55.6 & 56.7 & 48.6 & 48.2 & 48.3 & 47.1 & 52.2 \\
      & kNN-ICL \cite{good_examples_gpt3} & 55.7 & 56.2 & 56.6 & \textbf{57.3} & 57.0 & 51.1 & 51.6 & 50.8 & 51.6 & 51.5 \\
      & \modelname ($\alpha=1$) & 55.4 & 54.6 & 55.8 & 55.6 & 56.2 & 51.2 & 51.3 & \textbf{52.5} & \textbf{52.8} & 51.6 \\
      & \modelname ($\alpha=0.1$) & 55.5 & 55.1 & 55.4 & 57.0 & 55.8 & 49.0 & 50.1 & 51.9 & 52.6 & \textbf{53.3} \\
    \bottomrule
  \end{tabular}%
  }
  \caption{$F_{1}(\%)$ performance of Phi-4 and Llama8B with few-shot learning on three datasets. “Random” denotes standard prompting with randomly selected demonstration examples; \(\alpha = 1.0\) indicates that only hard examples are used; \(\alpha = 0.9\) corresponds to a composition of 90\% hard examples and 10\% kNN-ICL. Improvements shown in \textbf{bold} are statistically significant over 3 runs.}
  \label{tab:phi4_llama8}
\end{table*}


From the tables, we can see that across all LLMs, the Cognitive Distortion dataset proved the most challenging due to its closely related labels. Consequently, this dataset saw the largest performance gains from \modelname for both $\alpha=1.0$ and 0.9. This highlights the effectiveness of \modelname in handling label ambiguity, especially for challenging classification tasks.

From Table~\ref{tab:phi3_mistral7}, we can see that Phi-3-mini-8k excelled on Medical Abstracts, with \modelname ($\alpha=1.0$) yielding a 7.6\% $F_1$ improvement over random selection with 6-shot; $\alpha=0.9$ further improved upon kNN-ICL by 3\% at both 4-shot and 10-shot. Mistral-7B achieved optimal performance on Cognitive Distortion, surpassing random selection by 7.3\% at 10-shot. Also showed consistent gains on SST-5 and Medical Abstracts (3.5\% to 6.7\%). From Table~\ref{tab:phi4_llama8}, we can see that Phi4 showed maximal improvement on Cognitive Distortion with a 4.7\% increase. \modelname with $\alpha=0.9$ enhanced performance on Medical Abstracts. Llama3.1-8B achieved the best results on Cognitive Distortion and SST-5 (5.7\% $F_1$ gain).

\begin{figure*}[t]
    \centering
    \includegraphics[width=0.9\linewidth]{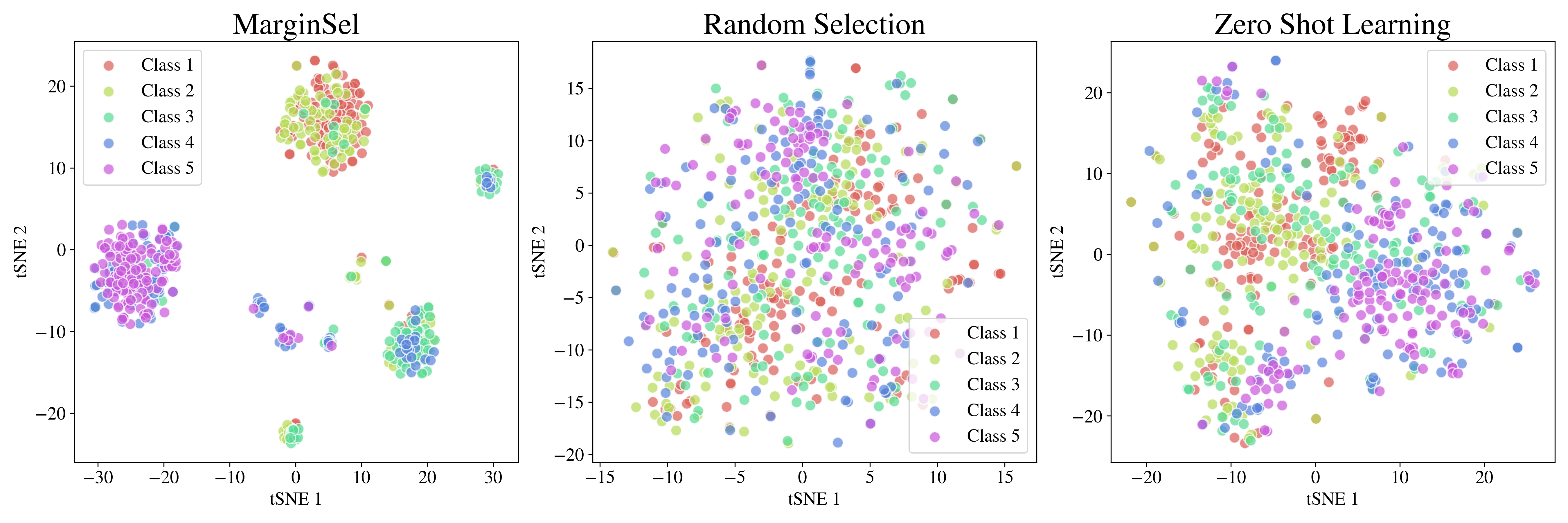}
    \caption{This figure shows the t-SNE\cite{tsne_paper} visualization of embeddings for each method. We can see that \modelname achieves superior class separation compared to others.}
    \label{fig:tsne_analysis}
\end{figure*}

However, we also observed that performance gains were not always consistent with increasing shot count, possibly due to context length limitations. For higher shot counts, kNN-ICL performance was often comparable to \modelname when $\alpha=1.0$, with $\alpha=0.9$ yielding marginal improvements in some cases. Performance varied across datasets and models, suggesting the influence of training data on model familiarity with the task. \modelname was particularly beneficial for models with smaller context windows, as careful example selection is more critical when the available token budget is limited.

Overall, we can conclude that \modelname is particularly useful for difficult classification tasks that contain ambiguity between labels and also for models with smaller context length.

\section{Analysis}

In this section, we analyze the impact of candidate labels and hard examples on the final prediction and classifier margin. More analysis on the no. of candidate labels assigned is in Section~\ref{sec:app:analysis}.

\begin{figure*}[t]
    \centering
    \includegraphics[width=0.95\linewidth]{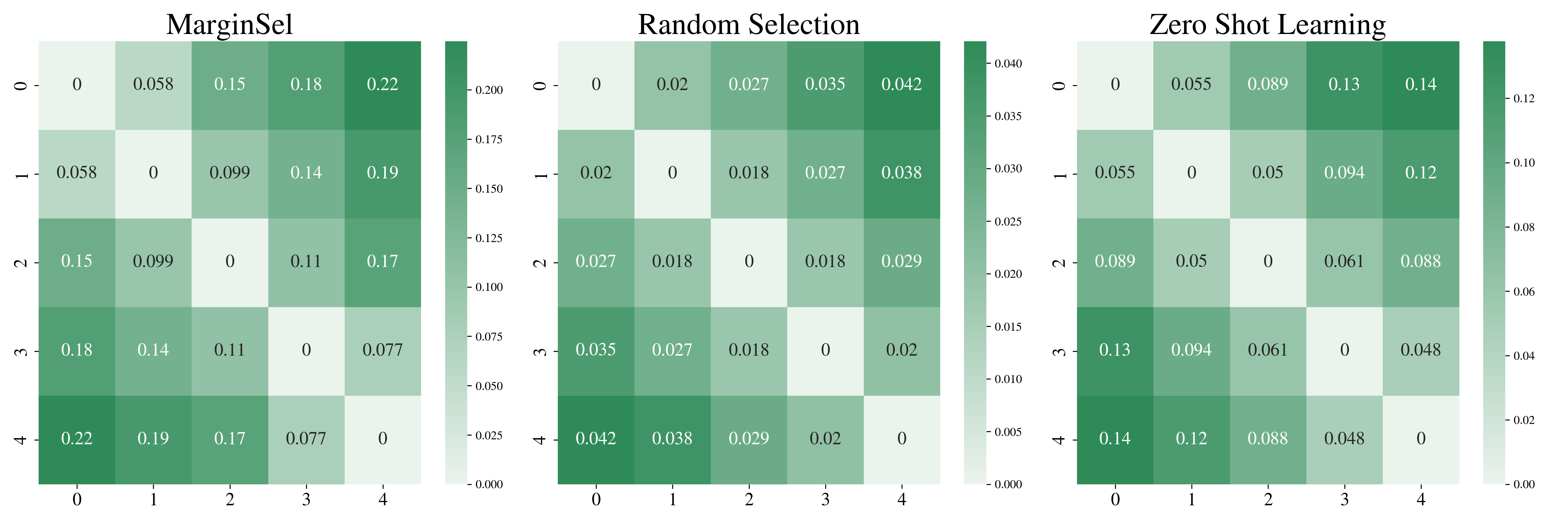}
    \caption{Heatmap of pairwise distances between classes for three methods: \modelname, Random Selection, and Zero-Shot Learning. The x-axis and y-axis represent the five classes from the SST-5 dataset. Notably, the inter-class distances are greatest for \modelname. For example, the distance between class-0 and class-4 is 0.22 for \modelname, compared to 0.042 for Random Selection and 0.14 for Zero-Shot Learning.}
    \label{fig:margin_analysis}
\end{figure*}

\subsection{Impact of Candidate Labels}
\label{sec:analysis_candidate_labels}
\begin{figure}[t]
    \centering
    \includegraphics[width=0.6\linewidth]{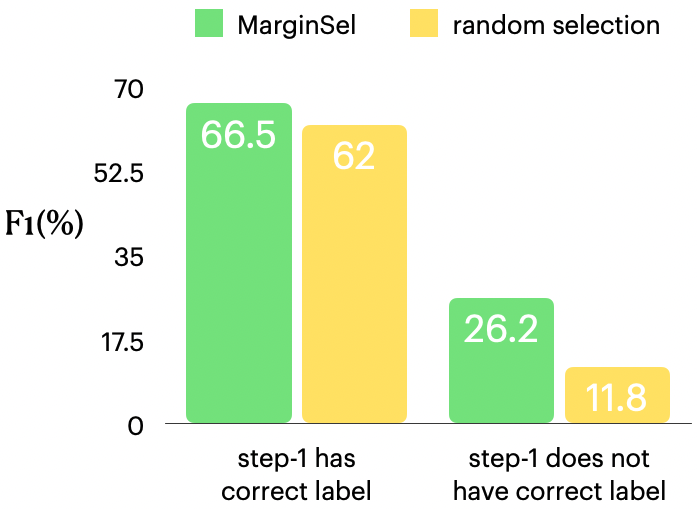}
    \caption{Impact of Step-1 on Overall Performance: The left bars represent the performance on examples for which the multiple outputs produced in Step-1 include the correct label, while the right bars show the performance on examples where the correct label is not present among the Step-1 outputs.}
    \label{fig:step1_analysis}
\end{figure}

\begin{figure}[t]
    \centering
    \includegraphics[width=0.9\linewidth]{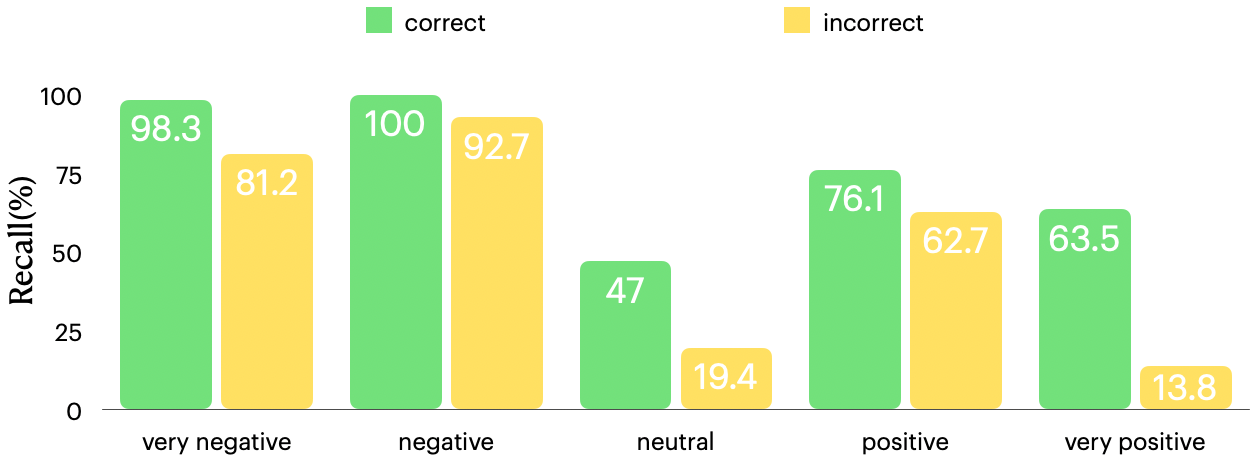}
    \caption{This figure shows the recall of step-1 when step-2 is correct/incorrect for 5 classes in the SST-5 dataset.}
    \label{fig:step2_analysis}
\end{figure}

In this experiment, we evaluate the impact of the Candidate Label Assignment step (Step-1). We conduct our evaluation with the Mistral-7B model on the SST-5 dataset, a less ambiguous dataset, to analyze \modelname's efficacy. Figure~\ref{fig:step1_analysis} shows the performance of \modelname and the baseline when the correct label is present (and absent) among the candidates labels obtained in Step 1. When the correct label is present, \modelname improves performance by 4.5\% over random selection. Even when the correct label is not present, a 14.4\% improvement is observed, indicating that Step-2 helps recover from Step-1 errors. This underscores the importance of Step-1, as discussed in Section~\ref{sec:hard_examples_help}.

Figure~\ref{fig:step2_analysis} further illustrates this observation. Here, we measure the recall of the correct label among the candidate labels (step-1) for instances where step-2 is correct(incorrect). It shows the recall of candidate labels in Step-1 relative to the final prediction across the five different labels in this dataset. We observe particularly high recall for the \emph{very negative}, \emph{negative}, and \emph{positive} labels when the final prediction is correct, further emphasizing the importance of accurately capturing relevant candidate labels in Step-1.

\subsection{Impact of Hard Examples on Classifier Margin}
\label{sec:analysis_margin}
We analyze the effect of hard examples on the classifier margin by examining embeddings from \modelname, random selection, and zero-shot learning. Analyzing the decision boundary directly is challenging due to its non-linear nature. However, in the embedding space, the decision boundary defined by the output layer remains constant. Thus, embedding movements relative to this layer indicate shifts in the overall decision boundary.

Figure~\ref{fig:margin_analysis} shows pairwise distances between class centroids computed from the embeddings of input prompts. The x and y axis show the five classes, and the value in the cell is the distance between each pair of classes. A darker color represents a high value of distance, indicating better separation between respective classes. We can see that \modelname exhibits greater inter-class separation than random and zero-shot methods. This corroborates the theory in Section~\ref{sec:other_examples_help} that hard examples maximize the margin. Figure~\ref{fig:tsne_analysis} visualizes the embeddings of various test instances using tSNE \cite{tsne_paper}, colored according to the labels of those instances. We can observe that different labels are better separated for \modelname than for the baselines.

\section{Conclusion}

In this work, we presented \modelname to dynamically select demonstration examples for in-context learning. By leveraging the concept of hard example mining, our approach identifies those examples that lie near the decision boundary—akin to support vectors in SVM—and uses them to effectively shift the classifier margin. Our theoretical analysis, inspired by the duality between transformer attention and gradient descent \cite{icl_meta_grad_theory}, establishes that \modelname can induce max-margin behavior in LLMs.

Extensive experiments on 3 diverse datasets across various text classification tasks (including sentiment analysis, abstract classification, and cognitive distortion detection) demonstrate that our method consistently outperforms strong baselines. Notably, we observe significant improvements in $F_{1}$ scores, with \modelname yielding gains of up to 7.6\%. Our margin analysis further confirms that the embeddings produced by \modelname lead to more clearly separated class clusters.

Together, the algorithm and theory underpin the benefits of \modelname for challenging text classification tasks. In future work, we plan to explore extensions of our method to other NLP tasks.

\clearpage

\section*{Limitations}
Although MarginSel shows promising improvements in handling ambiguous examples and inducing max-margin behavior, several limitations remain. First, the method relies on the quality of zero-shot candidate labels; if these initial predictions are unreliable, the effectiveness of the subsequent hard example selection may be compromised. Second, the selection ratio parameter \(\alpha\) is sensitive and may require careful, task-specific tuning. Third, the benefits of MarginSel are most pronounced in models with smaller context windows, which may limit its applicability to larger models with extensive context capacities. Although MarginSel avoids the cost of training a separate retriever, it still incurs computational overhead for generating candidate labels and selecting hard examples. This overhead might be significant for large datasets or complex LLMs. The theoretical analysis makes simplifying assumptions, such as using linear attention. These assumptions might not always hold in practice, potentially limiting the generalizability of the theoretical results. Finally, MarginSel has been primarily evaluated on text classification tasks, and its generalizability to other NLP applications remains to be explored.

\bibliography{references}

\begin{thebibliography}{18}
\providecommand{\natexlab}[1]{#1}

\bibitem[{Abdin et~al.(2024{\natexlab{a}})Abdin, Aneja, Awadalla, Awadallah, Awan, Bach, Bahree, Bakhtiari, Bao, Behl, Benhaim, Bilenko, Bjorck, Bubeck, Cai, Cai, Chaudhary, Chen, Chen, Chen, Chen, Chen, Cheng, Chopra, Dai, Dixon, Eldan, Fragoso, Gao, Gao, Gao, Garg, Giorno, Goswami, Gunasekar, Haider, Hao, Hewett, Hu, Huynh, Iter, Jacobs, Javaheripi, Jin, Karampatziakis, Kauffmann, Khademi, Kim, Kim, Kurilenko, Lee, Lee, Li, Li, Liang, Liden, Lin, Lin, Liu, Liu, Liu, Liu, Liu, Luo, Madan, Mahmoudzadeh, Majercak, Mazzola, Mendes, Mitra, Modi, Nguyen, Norick, Patra, Perez-Becker, Portet, Pryzant, Qin, Radmilac, Ren, de~Rosa, Rosset, Roy, Ruwase, Saarikivi, Saied, Salim, Santacroce, Shah, Shang, Sharma, Shen, Shukla, Song, Tanaka, Tupini, Vaddamanu, Wang, Wang, Wang, Wang, Wang, Wang, Ward, Wen, Witte, Wu, Wu, Wyatt, Xiao, Xu, Xu, Xu, Xue, Yadav, Yang, Yang, Yang, Yang, Yu, Yuan, Zhang, Zhang, Zhang, Zhang, Zhang, Zhang, Zhang, and Zhou}]{phi3_paper}
Marah Abdin, Jyoti Aneja, Hany Awadalla, Ahmed Awadallah, Ammar~Ahmad Awan, Nguyen Bach, Amit Bahree, Arash Bakhtiari, Jianmin Bao, Harkirat Behl, Alon Benhaim, Misha Bilenko, Johan Bjorck, Sébastien Bubeck, Martin Cai, Qin Cai, Vishrav Chaudhary, Dong Chen, Dongdong Chen, Weizhu Chen, Yen-Chun Chen, Yi-Ling Chen, Hao Cheng, Parul Chopra, Xiyang Dai, Matthew Dixon, Ronen Eldan, Victor Fragoso, Jianfeng Gao, Mei Gao, Min Gao, Amit Garg, Allie~Del Giorno, Abhishek Goswami, Suriya Gunasekar, Emman Haider, Junheng Hao, Russell~J. Hewett, Wenxiang Hu, Jamie Huynh, Dan Iter, Sam~Ade Jacobs, Mojan Javaheripi, Xin Jin, Nikos Karampatziakis, Piero Kauffmann, Mahoud Khademi, Dongwoo Kim, Young~Jin Kim, Lev Kurilenko, James~R. Lee, Yin~Tat Lee, Yuanzhi Li, Yunsheng Li, Chen Liang, Lars Liden, Xihui Lin, Zeqi Lin, Ce~Liu, Liyuan Liu, Mengchen Liu, Weishung Liu, Xiaodong Liu, Chong Luo, Piyush Madan, Ali Mahmoudzadeh, David Majercak, Matt Mazzola, Caio César~Teodoro Mendes, Arindam Mitra, Hardik Modi, Anh Nguyen,
  Brandon Norick, Barun Patra, Daniel Perez-Becker, Thomas Portet, Reid Pryzant, Heyang Qin, Marko Radmilac, Liliang Ren, Gustavo de~Rosa, Corby Rosset, Sambudha Roy, Olatunji Ruwase, Olli Saarikivi, Amin Saied, Adil Salim, Michael Santacroce, Shital Shah, Ning Shang, Hiteshi Sharma, Yelong Shen, Swadheen Shukla, Xia Song, Masahiro Tanaka, Andrea Tupini, Praneetha Vaddamanu, Chunyu Wang, Guanhua Wang, Lijuan Wang, Shuohang Wang, Xin Wang, Yu~Wang, Rachel Ward, Wen Wen, Philipp Witte, Haiping Wu, Xiaoxia Wu, Michael Wyatt, Bin Xiao, Can Xu, Jiahang Xu, Weijian Xu, Jilong Xue, Sonali Yadav, Fan Yang, Jianwei Yang, Yifan Yang, Ziyi Yang, Donghan Yu, Lu~Yuan, Chenruidong Zhang, Cyril Zhang, Jianwen Zhang, Li~Lyna Zhang, Yi~Zhang, Yue Zhang, Yunan Zhang, and Xiren Zhou. 2024{\natexlab{a}}.
\newblock \href {https://arxiv.org/abs/2404.14219} {Phi-3 technical report: A highly capable language model locally on your phone}.
\newblock \emph{Preprint}, arXiv:2404.14219.

\bibitem[{Abdin et~al.(2024{\natexlab{b}})Abdin, Aneja, Behl, Bubeck, Eldan, Gunasekar, Harrison, Hewett, Javaheripi, Kauffmann, Lee, Lee, Li, Liu, Mendes, Nguyen, Price, de~Rosa, Saarikivi, Salim, Shah, Wang, Ward, Wu, Yu, Zhang, and Zhang}]{phi4_paper}
Marah Abdin, Jyoti Aneja, Harkirat Behl, Sébastien Bubeck, Ronen Eldan, Suriya Gunasekar, Michael Harrison, Russell~J. Hewett, Mojan Javaheripi, Piero Kauffmann, James~R. Lee, Yin~Tat Lee, Yuanzhi Li, Weishung Liu, Caio C.~T. Mendes, Anh Nguyen, Eric Price, Gustavo de~Rosa, Olli Saarikivi, Adil Salim, Shital Shah, Xin Wang, Rachel Ward, Yue Wu, Dingli Yu, Cyril Zhang, and Yi~Zhang. 2024{\natexlab{b}}.
\newblock \href {https://arxiv.org/abs/2412.08905} {Phi-4 technical report}.
\newblock \emph{Preprint}, arXiv:2412.08905.

\bibitem[{Brown et~al.(2020)Brown, Mann, Ryder, Subbiah, Kaplan, Dhariwal, Neelakantan, Shyam, Sastry, Askell, Agarwal, Herbert-Voss, Krueger, Henighan, Child, Ramesh, Ziegler, Wu, Winter, Hesse, Chen, Sigler, Litwin, Gray, Chess, Clark, Berner, McCandlish, Radford, Sutskever, and Amodei}]{gpt3_paper}
Tom Brown, Benjamin Mann, Nick Ryder, Melanie Subbiah, Jared~D Kaplan, Prafulla Dhariwal, Arvind Neelakantan, Pranav Shyam, Girish Sastry, Amanda Askell, Sandhini Agarwal, Ariel Herbert-Voss, Gretchen Krueger, Tom Henighan, Rewon Child, Aditya Ramesh, Daniel Ziegler, Jeffrey Wu, Clemens Winter, Chris Hesse, Mark Chen, Eric Sigler, Mateusz Litwin, Scott Gray, Benjamin Chess, Jack Clark, Christopher Berner, Sam McCandlish, Alec Radford, Ilya Sutskever, and Dario Amodei. 2020.
\newblock \href {https://proceedings.neurips.cc/paper_files/paper/2020/file/1457c0d6bfcb4967418bfb8ac142f64a-Paper.pdf} {Language models are few-shot learners}.
\newblock In \emph{Advances in Neural Information Processing Systems}, volume~33, pages 1877--1901. Curran Associates, Inc.

\bibitem[{Dai et~al.(2023)Dai, Sun, Dong, Hao, Ma, Sui, and Wei}]{icl_meta_grad_theory}
Damai Dai, Yutao Sun, Li~Dong, Yaru Hao, Shuming Ma, Zhifang Sui, and Furu Wei. 2023.
\newblock \href {https://doi.org/10.18653/v1/2023.findings-acl.247} {Why can {GPT} learn in-context? language models secretly perform gradient descent as meta-optimizers}.
\newblock In \emph{Findings of the Association for Computational Linguistics: ACL 2023}, pages 4005--4019, Toronto, Canada. Association for Computational Linguistics.

\bibitem[{Jiang et~al.(2023)Jiang, Sablayrolles, Mensch, Bamford, Chaplot, de~las Casas, Bressand, Lengyel, Lample, Saulnier, Lavaud, Lachaux, Stock, Scao, Lavril, Wang, Lacroix, and Sayed}]{mistral_paper}
Albert~Q. Jiang, Alexandre Sablayrolles, Arthur Mensch, Chris Bamford, Devendra~Singh Chaplot, Diego de~las Casas, Florian Bressand, Gianna Lengyel, Guillaume Lample, Lucile Saulnier, Lélio~Renard Lavaud, Marie-Anne Lachaux, Pierre Stock, Teven~Le Scao, Thibaut Lavril, Thomas Wang, Timothée Lacroix, and William~El Sayed. 2023.
\newblock \href {https://arxiv.org/abs/2310.06825} {Mistral 7b}.
\newblock \emph{Preprint}, arXiv:2310.06825.

\bibitem[{Levy et~al.(2023)Levy, Bogin, and Berant}]{levy2023diversedemonstrationsimproveincontext}
Itay Levy, Ben Bogin, and Jonathan Berant. 2023.
\newblock \href {https://arxiv.org/abs/2212.06800} {Diverse demonstrations improve in-context compositional generalization}.
\newblock \emph{Preprint}, arXiv:2212.06800.

\bibitem[{Liu et~al.(2022)Liu, Shen, Zhang, Dolan, Carin, and Chen}]{good_examples_gpt3}
Jiachang Liu, Dinghan Shen, Yizhe Zhang, Bill Dolan, Lawrence Carin, and Weizhu Chen. 2022.
\newblock \href {https://doi.org/10.18653/v1/2022.deelio-1.10} {What makes good in-context examples for {GPT}-3?}
\newblock In \emph{Proceedings of Deep Learning Inside Out (DeeLIO 2022): The 3rd Workshop on Knowledge Extraction and Integration for Deep Learning Architectures}, pages 100--114, Dublin, Ireland and Online. Association for Computational Linguistics.

\bibitem[{Liu et~al.(2020)Liu, Ott, Goyal, Du, Joshi, Chen, Levy, Lewis, Zettlemoyer, and Stoyanov}]{liu2020roberta}
Yinhan Liu, Myle Ott, Naman Goyal, Jingfei Du, Mandar Joshi, Danqi Chen, Omer Levy, Mike Lewis, Luke Zettlemoyer, and Veselin Stoyanov. 2020.
\newblock \href {https://openreview.net/forum?id=SyxS0T4tvS} {Ro{\{}bert{\}}a: A robustly optimized {\{}bert{\}} pretraining approach}.

\bibitem[{Lu et~al.(2023)Lu, Qiu, Chang, Wu, Zhu, Rajpurohit, Clark, and Kalyan}]{lu2023dynamicpromptlearningpolicyPGPrompt}
Pan Lu, Liang Qiu, Kai-Wei Chang, Ying~Nian Wu, Song-Chun Zhu, Tanmay Rajpurohit, Peter Clark, and Ashwin Kalyan. 2023.
\newblock \href {https://arxiv.org/abs/2209.14610} {Dynamic prompt learning via policy gradient for semi-structured mathematical reasoning}.
\newblock \emph{Preprint}, arXiv:2209.14610.

\bibitem[{Lu et~al.(2022)Lu, Bartolo, Moore, Riedel, and Stenetorp}]{order_sensitivity}
Yao Lu, Max Bartolo, Alastair Moore, Sebastian Riedel, and Pontus Stenetorp. 2022.
\newblock \href {https://doi.org/10.18653/v1/2022.acl-long.556} {Fantastically ordered prompts and where to find them: Overcoming few-shot prompt order sensitivity}.
\newblock In \emph{Proceedings of the 60th Annual Meeting of the Association for Computational Linguistics (Volume 1: Long Papers)}, pages 8086--8098, Dublin, Ireland. Association for Computational Linguistics.

\bibitem[{Rubin et~al.(2022)Rubin, Herzig, and Berant}]{rubin-etal-2022-retriever}
Ohad Rubin, Jonathan Herzig, and Jonathan Berant. 2022.
\newblock \href {https://doi.org/10.18653/v1/2022.naacl-main.191} {Learning to retrieve prompts for in-context learning}.
\newblock In \emph{Proceedings of the 2022 Conference of the North American Chapter of the Association for Computational Linguistics: Human Language Technologies}, pages 2655--2671, Seattle, United States. Association for Computational Linguistics.

\bibitem[{Schopf et~al.(2023)Schopf, Braun, and Matthes}]{medabs_dataset}
Tim Schopf, Daniel Braun, and Florian Matthes. 2023.
\newblock \href {https://doi.org/10.1145/3582768.3582795} {Evaluating unsupervised text classification: Zero-shot and similarity-based approaches}.
\newblock In \emph{Proceedings of the 2022 6th International Conference on Natural Language Processing and Information Retrieval}, NLPIR '22, page 6–15, New York, NY, USA. Association for Computing Machinery.

\bibitem[{Shreevastava and Foltz(2021)}]{cogdist_dataset}
Sagarika Shreevastava and Peter Foltz. 2021.
\newblock \href {https://doi.org/10.18653/v1/2021.clpsych-1.17} {Detecting cognitive distortions from patient-therapist interactions}.
\newblock In \emph{Proceedings of the Seventh Workshop on Computational Linguistics and Clinical Psychology: Improving Access}, pages 151--158, Online. Association for Computational Linguistics.

\bibitem[{Shrivastava et~al.(2016)Shrivastava, Gupta, and Girshick}]{hard_mining}
Abhinav Shrivastava, Abhinav Gupta, and Ross~B. Girshick. 2016.
\newblock \href {https://arxiv.org/abs/1604.03540} {Training region-based object detectors with online hard example mining}.
\newblock \emph{CoRR}, abs/1604.03540.

\bibitem[{Socher et~al.(2013)Socher, Perelygin, Wu, Chuang, Manning, Ng, and Potts}]{sst_dataset}
Richard Socher, Alex Perelygin, Jean Wu, Jason Chuang, Christopher~D. Manning, Andrew Ng, and Christopher Potts. 2013.
\newblock \href {https://aclanthology.org/D13-1170/} {Recursive deep models for semantic compositionality over a sentiment treebank}.
\newblock In \emph{Proceedings of the 2013 Conference on Empirical Methods in Natural Language Processing}, pages 1631--1642, Seattle, Washington, USA. Association for Computational Linguistics.

\bibitem[{Touvron et~al.(2023)Touvron, Lavril, Izacard, Martinet, Lachaux, Lacroix, Rozière, Goyal, Hambro, Azhar, Rodriguez, Joulin, Grave, and Lample}]{llama_paper}
Hugo Touvron, Thibaut Lavril, Gautier Izacard, Xavier Martinet, Marie-Anne Lachaux, Timothée Lacroix, Baptiste Rozière, Naman Goyal, Eric Hambro, Faisal Azhar, Aurelien Rodriguez, Armand Joulin, Edouard Grave, and Guillaume Lample. 2023.
\newblock \href {https://arxiv.org/abs/2302.13971} {Llama: Open and efficient foundation language models}.
\newblock \emph{Preprint}, arXiv:2302.13971.

\bibitem[{van~der Maaten and Hinton(2008)}]{tsne_paper}
Laurens van~der Maaten and Geoffrey Hinton. 2008.
\newblock \href {http://jmlr.org/papers/v9/vandermaaten08a.html} {Visualizing data using t-sne}.
\newblock \emph{Journal of Machine Learning Research}, 9(86):2579--2605.

\bibitem[{Wang et~al.(2024)Wang, Wu, Yuan, Li, Cai, and Jia}]{wang2024demonstrationselectionincontextlearningRDES}
Xubin Wang, Jianfei Wu, Yichen Yuan, Mingzhe Li, Deyu Cai, and Weijia Jia. 2024.
\newblock \href {https://arxiv.org/abs/2412.03966} {Demonstration selection for in-context learning via reinforcement learning}.
\newblock \emph{Preprint}, arXiv:2412.03966.

\end{thebibliography}

\appendix
\section{Experiments}

\subsection{Dataset Details}
\label{app:datasets}
This appendix provides detailed descriptions of the datasets used in our evaluations.

\subsubsection{SST-5}
The Stanford Sentiment Treebank \cite{sst_dataset} contains 11,855 phrases annotated with five sentiment labels: negative, somewhat negative, neutral, somewhat positive, and positive. For our evaluations, we constructed stratified splits comprising 850 training examples and 730 test examples.

\subsubsection{Cognitive Distortion}
The Cognitive Distortion dataset is derived from 2,530 patient responses in the Therapist Q\&A dataset\footnote{\url{https://www.kaggle.com/datasets/arnmaud/therapist-qa}}, as annotated by \citet{cogdist_dataset}. Each response is labeled with one of 10 possible cognitive distortions. For our evaluations, we focus on 5 specific cognitive distortions: emotional reasoning, mental filter, overgeneralization, personalization, and mind reading. We performed stratified sampling to obtain 443 examples each for training and testing.

\subsubsection{Medical Abstracts}
The Medical Abstracts dataset was constructed by \citet{medabs_dataset} using raw medical abstracts from Kaggle\footnote{\url{https://www.kaggle.com/datasets/chaitanyakck/medical-text}}. The dataset comprises 11,600 training examples and 2,890 test examples, each annotated with one of the following descriptive labels: neoplasms, digestive system diseases, nervous system diseases, cardiovascular diseases, and general pathological conditions. For our evaluations, we performed stratified sampling to obtain 1,050 training samples and 722 test samples.

\subsection{Prompts}
\label{sec:prompts}
The system and user prompts for the candidate label generation step, as well as the final label prediction (ICL), for all datasets are shown in Figures~\ref{prompt:cogdist_step1}--\ref{prompt:sst5_step3}. We use the same prompts for all four models.

\begin{figure*}[ht]
  \centering
    \begin{tcolorbox}[title={System Prompt}, colback=blue!5, colframe=blue!75!black, drop shadow={shadow xshift=1mm, shadow yshift=1mm, opacity=0.25}, width=\textwidth]
    You are an expert in cognitive distortion detection. Your goal is to assign label(s) to each text based on the type of cognitive distortion present:\\
    - mental filter: When the text focuses exclusively on negative details while ignoring positive ones.\\
    - overgeneralization: When the text sees a single negative event as a never-ending pattern.\\
    - personalization: When the text blames oneself for events outside one's control.\\
    - emotional reasoning: When the text assumes that negative emotions reflect reality.\\
    - mind reading: When the text assumes what others are thinking without evidence.
    \end{tcolorbox}
    \begin{tcolorbox}[title={User Prompt}, colback=green!5, colframe=green!75!black, drop shadow={shadow xshift=1mm, shadow yshift=1mm, opacity=0.25}, width=\textwidth]
    Given the text: '\{ \}', you must carefully analyze EVERY POSSIBLE cognitive distortion present. 
    For EACH label below, independently consider if it applies (even partially) to the text:\\
    - mental filter: Does the text focus exclusively on negative details while ignoring positive ones? (even partially)\\
    - overgeneralization: Does the text see a single negative event as a never-ending pattern? (even partially)\\
    - personalization: Does the text blame oneself for events outside one's control? (even partially)\\
    - emotional reasoning: Does the text assume that negative emotions reflect reality? (even partially)\\
    - mind reading: Does the text assume what others are thinking without evidence? (even partially)\\

    IMPORTANT:\\
    - Evaluate each label separately - the presence of one label doesn't exclude others\\
    - Even slight or partial matches should be included\\
    - Many texts may exhibit 1+ distortions simultaneously\\
    - When in doubt, include the label\\

    Return ALL relevant labels in comma-separated format within the <label></label> tags (e.g., <label>mental filter,overgeneralization,personalization,emotional reasoning,mind reading</label>).
    \end{tcolorbox}    
  \caption{Prompts used for the \textit{Candidate Label Assignment} step for the Cognitive Distortion Dataset\cite{cogdist_dataset}.}
  \label{prompt:cogdist_step1}
\end{figure*}

\begin{figure*}[ht]
  \centering
    \begin{tcolorbox}[title={System Prompt}, colback=blue!5, colframe=blue!75!black, drop shadow={shadow xshift=1mm, shadow yshift=1mm, opacity=0.25}, width=\textwidth]
        You are an expert in cognitive distortion detection. Your goal is to assign each text a label based on the type of cognitive distortion present:\\
        - mental filter: When the text focuses exclusively on negative details while ignoring positive ones.\\
        - overgeneralization: When the text sees a single negative event as a never-ending pattern.\\
        - personalization: When the text blames oneself for events outside one's control.\\
        - emotional reasoning: When the text assumes that negative emotions reflect reality.\\
        - mind reading: When the text assumes what others are thinking without evidence.
    \end{tcolorbox}
    \begin{tcolorbox}[title={User Prompt}, colback=green!5, colframe=green!75!black,  drop shadow={shadow xshift=1mm, shadow yshift=1mm, opacity=0.25},  width=\textwidth]
    Given the text: '\{ \}', analyze the cognitive distortion present step-by-step. 
    Identify which cognitive distortion label is most appropriate based on content, tone, and context. 
    Provide the label exactly as follows: <label>label</label>, where 'label' is one of the following:
    - mental filter\\
    - overgeneralization\\
    - personalization\\
    - emotional reasoning\\
    - mind reading\\
    Do not include any additional formatting or characters, just return the label within the <label></label> tags.
    \end{tcolorbox}    
  \caption{Prompts used for the \textit{Final Label Consolidation} step for the Cognitive Distortion Dataset\cite{cogdist_dataset}.}
  \label{prompt:cogdist_step3}
\end{figure*}

\begin{figure*}[ht]
  \centering
    \begin{tcolorbox}[title={System Prompt}, colback=blue!5, colframe=blue!75!black, drop shadow={shadow xshift=1mm, shadow yshift=1mm, opacity=0.25}, width=\textwidth]
    You are an expert in medical text analysis. Your goal is to assign label(s) to each medical abstract based on its content:\\
    - neoplasms: Abstracts related to tumors, cancers, or abnormal tissue growth.\\
    - digestive system diseases: Abstracts related to diseases of the digestive system, such as Crohn's disease or ulcers.\\
    - nervous system diseases: Abstracts related to diseases of the nervous system, such as Alzheimer's or Parkinson's disease.\\
    - cardiovascular diseases: Abstracts related to diseases of the heart and blood vessels, such as hypertension or heart failure.\\
    - general pathological conditions: Abstracts related to general pathological conditions, such as inflammation or infection.
    \end{tcolorbox}
    
    \begin{tcolorbox}[title={User Prompt}, colback=green!5, colframe=green!75!black,  drop shadow={shadow xshift=1mm, shadow yshift=1mm, opacity=0.25}, width=\textwidth]
    Given the medical abstract: '\{ \}', you must carefully analyze EVERY POSSIBLE topic expressed in the abstract.
    For EACH label below, independently consider if it applies (even partially) to the abstract:\\
    - neoplasms: Does the abstract discuss ANY topics related to tumors, cancers, or abnormal tissue growth?\\
    - digestive system diseases: Does the abstract discuss ANY topics related to diseases of the digestive system, such as Crohn's disease or ulcers?\\
    - nervous system diseases: Does the abstract discuss ANY topics related to diseases of the nervous system, such as Alzheimer's or Parkinson's disease?\\
    - cardiovascular diseases: Does the abstract discuss ANY topics related to diseases of the heart and blood vessels, such as hypertension or heart failure?\\
    - general pathological conditions: Does the abstract discuss ANY topics related to general pathological conditions, such as inflammation or infection?\\

    IMPORTANT:\\
    - Evaluate each label separately—the presence of one label doesn't exclude others.\\
    - Even slight or partial matches should be included.\\
    - Abstracts can discuss multiple topics simultaneously.\\
    - When in doubt, include the label.\\

    Return ALL relevant labels in comma-separated format within the <label></label> tags (e.g., <label>neoplasms,digestive system diseases,nervous system diseases,cardiovascular diseases,general pathological conditions</label>).
    \end{tcolorbox}    
  \caption{Prompts used for the \textit{Candidate Label Assignment} step for the Medical Abstracts Dataset\cite{medabs_dataset}.}
  \label{prompt:medabs_step1}
\end{figure*}

\begin{figure*}[ht]
  \centering
    \begin{tcolorbox}[title={System Prompt}, colback=blue!5, colframe=blue!75!black, drop shadow={shadow xshift=1mm, shadow yshift=1mm, opacity=0.25}, width=\textwidth]
    You are an expert in medical text analysis. Your goal is to assign each medical abstract a label based on its content:\\
    - neoplasms: Abstracts related to tumors, cancers, or abnormal tissue growth.\\
    - digestive system diseases: Abstracts related to diseases of the digestive system, such as Crohn's disease or ulcers.\\
    - nervous system diseases: Abstracts related to diseases of the nervous system, such as Alzheimer's or Parkinson's disease.\\
    - cardiovascular diseases: Abstracts related to diseases of the heart and blood vessels, such as hypertension or heart failure.\\
    - general pathological conditions: Abstracts related to general pathological conditions, such as inflammation or infection.
    \end{tcolorbox}
    \begin{tcolorbox}[title={User Prompt}, colback=green!5, colframe=green!75!black,  drop shadow={shadow xshift=1mm, shadow yshift=1mm, opacity=0.25},  width=\textwidth]
    Given the medical abstract: '\{ \}', analyze the content step-by-step.
    Identify which field of study label is most appropriate based on the topic, methodology, and context.
    Provide the label exactly as follows: <label>label</label>, where 'label' is one of the following:\\
    - neoplasms\\
    - digestive system diseases\\
    - nervous system diseases\\
    - cardiovascular diseases\\
    - general pathological conditions\\
    Do not include any additional formatting or characters, just return the label within the <label></label> tags.
    \end{tcolorbox}    
  \caption{Prompts used for the \textit{Final Label Consolidation} step for the Medical Abstracts Dataset\cite{medabs_dataset}.}
  \label{prompt:medabs_step3}
\end{figure*}

\begin{figure*}[ht]
  \centering
    \begin{tcolorbox}[title={System Prompt}, colback=blue!5, colframe=blue!75!black, drop shadow={shadow xshift=1mm, shadow yshift=1mm, opacity=0.25}, width=\textwidth]
    You are an expert in sentiment analysis of movie reviews. Your goal is to assign label(s) to each review based on its sentiment:\\
    - very negative: Reviews that express extremely unfavorable opinions, strong criticism, or intense dissatisfaction regarding the movie.\\
    - negative: Reviews that express unfavorable opinions, criticism, or dissatisfaction regarding the movie.\\
    - neutral: Reviews that express neither strong positive nor strong negative opinions, or that are balanced between praise and criticism.\\
    - positive: Reviews that express favorable opinions, praise, or satisfaction regarding the movie.\\
    - very positive: Reviews that express extremely favorable opinions, strong praise, or intense satisfaction regarding the movie.
    \end{tcolorbox}
    
    \begin{tcolorbox}[title={User Prompt}, colback=green!5, colframe=green!75!black,  drop shadow={shadow xshift=1mm, shadow yshift=1mm, opacity=0.25},  width=\textwidth]
    Given the movie review: '\{ \}', you must carefully analyze EVERY POSSIBLE sentiment expressed in the review.
    For EACH label below, independently consider if it applies (even partially) to the review:\\
    - very negative: Does the review express ANY extremely unfavorable opinions, strong criticism, or intense dissatisfaction regarding the movie?\\
    - negative: Does the review express ANY unfavorable opinions, criticism, or dissatisfaction regarding the movie?\\
    - neutral: Does the review express ANY neutral opinions, or is it balanced between praise and criticism?\\
    - positive: Does the review express ANY favorable opinions, praise, or satisfaction regarding the movie?\\
    - very positive: Does the review express ANY extremely favorable opinions, strong praise, or intense satisfaction regarding the movie?\\

    IMPORTANT:\\
    - Evaluate each label separately—the presence of one label doesn't exclude others.\\
    - Even slight or partial matches should be included.\\
    - Reviews can express mixed sentiments.\\
    - When in doubt, include the label.\\

    Return ALL relevant labels in comma-separated format within the <label></label> tags (e.g., <label>very negative,negative,neutral,positive,very positive</label>).
    \end{tcolorbox}    
  \caption{Prompts used for the \textit{Candidate Label Assignment} step for the SST-5 Dataset\cite{sst_dataset}.}
  \label{prompt:sst5_step1}
\end{figure*}

\begin{figure*}[ht]
  \centering
    \begin{tcolorbox}[title={System Prompt}, colback=blue!5, colframe=blue!75!black, drop shadow={shadow xshift=1mm, shadow yshift=1mm, opacity=0.25}, width=\textwidth]
    You are an expert in sentiment analysis of movie reviews. Your goal is to assign each review a label based on its sentiment:\\
    - very negative: Reviews that express extremely unfavorable opinions, strong criticism, or intense dissatisfaction regarding the movie.\\
    - negative: Reviews that express unfavorable opinions, criticism, or dissatisfaction regarding the movie.\\
    - neutral: Reviews that express neither strong positive nor strong negative opinions, or that are balanced between praise and criticism.\\
    - positive: Reviews that express favorable opinions, praise, or satisfaction regarding the movie.\\
    - very positive: Reviews that express extremely favorable opinions, strong praise, or intense satisfaction regarding the movie.
    \end{tcolorbox}
    \begin{tcolorbox}[title={User Prompt}, colback=green!5, colframe=green!75!black,  drop shadow={shadow xshift=1mm, shadow yshift=1mm, opacity=0.25},  width=\textwidth]
    Given the movie review: '\{ \}', analyze the sentiment expressed in the review step-by-step.
    Identify which sentiment label is most appropriate based on content, tone, and context.
    Provide the label exactly as follows: <label>label</label>, where 'label' is one of the following:\\
    - very negative\\
    - negative\\
    - neutral\\
    - positive\\
    - very positive\\

    Do not include any additional formatting or characters, just return the label within the <label></label> tags.
    \end{tcolorbox}    
  \caption{Prompts used for the \textit{Final Label Consolidation} step for the SST-5 Dataset\cite{sst_dataset}.}
  \label{prompt:sst5_step3}
\end{figure*}

\subsection{Results}
\label{sec:app:results}
\begin{figure*}[t]
    \centering
    \includegraphics[width=0.6\linewidth]{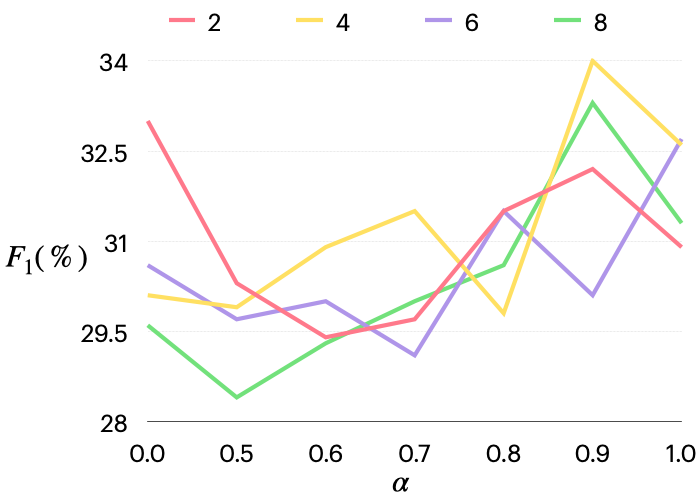}
    \caption{$F_{1}$ score of Phi3-mini-8k on the Cognitive Distortion dataset for different values of $\alpha$ and shot count. Here, $\alpha$ denotes the ratio of the number of \modelname samples to the number of kNN-ICL samples.}
    \label{fig:alpha_tuning}
\end{figure*}

For hyperparameter tuning of $\alpha$ on the Cognitive Distortion dataset, we performed stratified sampling to create a validation set of 221 examples. As discussed in the Section~\ref{sec:method}, $\alpha$ is defined as the ratio of the number of \modelname\ examples to the number of kNN-ICL examples included in the ICL prompt. Figure~\ref{fig:alpha_tuning} shows the performance of \modelname\ for different values of $\alpha$. We observe that performance peaks at $\alpha=0.9$ and deteriorates as more kNN-ICL samples (smaller $\alpha)$ are added, suggesting that the majority of examples should come from \modelname\ for optimal results. This finding corroborates the theory in Section~\ref{sec:other_examples_help} that incorporating additional examples does not necessarily enhance performance. The modest improvement from including 10\% of kNN-ICL examples can be attributed to errors in step 1 of \modelname, as discussed in Section~\ref{sec:analysis_candidate_labels}. Furthermore, adding more kNN-ICL examples may introduce noise into the second term of Equation~\ref{eq:icl_zsl_sum}, thereby deviating the query representation further from the true label representation in the third term.

\begin{table*}[ht]
  \centering
  {\small
  \begin{tabularx}{0.8\textwidth}{ll*{5}{>{\centering\arraybackslash}X}}
    \toprule
    \multirow{2}{*}{Dataset} & \multirow{2}{*}{Method} & \multicolumn{5}{c}{Few-Shot} \\
    \cmidrule(lr){3-7}
    & & 2& 4& 6& 8& 10\\
    \midrule
    \multirow{4}{*}{\makecell{Cognitive \\ Distortion}} & Random & 37.4 & 38.9 & 36.5 & 34.6 & 33.5 \\
                               & kNN-ICL \cite{good_examples_gpt3} & 39.9 & 39.4 & 39.2 & 38.8 & \textbf{40.1} \\
                               & \modelname ($\alpha=1$) & 38.0 & 39.8 & 37.8 & 37.6 & 36.9 \\
                               & \modelname ($\alpha=0.9$)  & 36.9 & 37.0 & 37.0 & 38.0 & 37.8 \\
    \midrule
    \multirow{4}{*}{\makecell{Medical \\ Abstraction}} & Random   & 64.1 & 64.5 & 63.9 & 63.6 & 60.8 \\
                               & kNN-ICL \cite{good_examples_gpt3}  & 65.2 & 64.8 & 64.7 & 64.9 & 64.0 \\
                               & \modelname ($\alpha=1$)  & 64.2 & 63.9 & 63.7 & 63.3 & 63.7 \\
                               & \modelname ($\alpha=0.9$)  & 64.2 & 64.2 & \textbf{65.0} & 64.4 & 64.5 \\
    \midrule
    \multirow{4}{*}{SST-5} & Random & 57.0 & 58.2 & 57.0 & 58.3 & 56.8 \\
                               & kNN-ICL \cite{good_examples_gpt3}  & 56.8 & 57.2 & 57.4 & 57.7 & 56.1 \\
                               & \modelname ($\alpha=1$)  & 57.0 & 58.0 & 57.1 & 57.6 & \textbf{57.7} \\
                               & \modelname ($\alpha=0.9$)  & 58.0 & 55.9 & 56.6 & 56.6 & 56.4 \\
    \bottomrule
  \end{tabularx}
  }
  \caption{$F_{1}(\%)$ performance of Llama-3.1-70B (8-bit quantization) with few-shot learning on three datasets. “Random” denotes standard prompting with randomly selected demonstration examples; \(\alpha = 1.0\) indicates that only hard examples are used; \(\alpha = 0.9\) corresponds to a composition of 90\% hard examples and 10\% kNN-ICL. Improvements shown in \textbf{bold} are statistically significant over 3 runs.}
    \label{tab:llama70_results}
\end{table*}

\subsection{Analysis}
\label{sec:app:analysis}
\begin{figure*}[t]
    \centering
    \includegraphics[width=\linewidth]{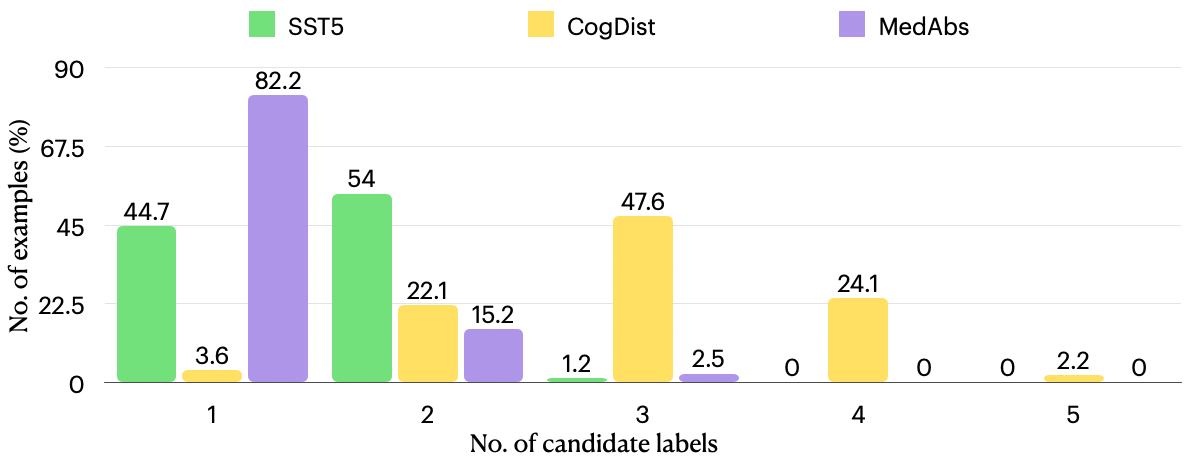}
    \caption{Histogram showing the number of candidate labels assigned in Step-1 across the three datasets.}
    \label{fig:candidate_label_analysis}
\end{figure*}

We also analyzed the number of candidate labels assigned in Step-1. Figure~\ref{fig:candidate_label_analysis} shows a histogram of the candidate labels assigned across instances for each dataset. For our experiments, we used Mistral-7B with 10 shots. From the figure, we observe that only one candidate label is assigned to 82.2\% of the examples in the Medical Abstraction dataset, making it the least ambiguous dataset. In contrast, 96.4\% of the examples in the Cognitive Distortion dataset are assigned more than one candidate label, with three candidate labels being the most common. This indicates that the Cognitive Distortion dataset is the most ambiguous dataset, consistent with the largest performance gains of \modelname observed in Section~\ref{sec:results}. This further reinforces the efficacy of \modelname for difficult classification tasks that contain ambiguity between labels.

\end{document}